\documentclass[sigconf]{acmart}

\usepackage{graphicx}
\usepackage{tabularray}

\AtBeginDocument{%
  \providecommand\BibTeX{{%
    \normalfont B\kern-0.5em{\scshape i\kern-0.25em b}\kern-0.8em\TeX}}}

\acmConference[FIRE '23]{Forum for Information Retrieval Evaluation}{December 15--18, 2023}{Goa University, Panjim}
\copyrightyear{2023}
\acmYear{2023}
\setcopyright{rightsretained}
\acmConference[FIRE 2023]{Forum for Information Retrieval
Evaluation}{December 15--18, 2023}{Panjim, India}
\acmBooktitle{Forum for Information Retrieval Evaluation (FIRE 2023), December
15--18, 2023, Panjim, India}\acmDOI{10.1145/3632754.3632765}
\acmISBN{979-8-4007-1632-4/23/12}

\begin{document}




\title{Fact-based Court Judgment Prediction}

\author{Shubham Kumar Nigam}
\email{shubhamkumarnigam@gmail.com}
\affiliation{%
  \institution{Indian Institute of Technology}
  \city{Kanpur}
  \country{India}
}

\author{Aniket Deroy}
\email{roydanik18@kgpian.iitkgp.ac.in}
\affiliation{%
  \institution{Indian Institute of Technology}
  \city{Kharagpur}
  \country{India}
}


\begin{abstract}

This extended abstract extends the research presented in "ILDC for CJPE: Indian Legal Documents Corpus for Court Judgment Prediction and Explanation" \cite{malik-etal-2021-ildc}, focusing on fact-based judgment prediction within the context of Indian legal documents. We introduce two distinct problem variations: one based solely on facts, and another combining facts with rulings from lower courts (RLC). Our research aims to enhance early-phase case outcome prediction, offering significant benefits to legal professionals and the general public. The results, however, indicated a performance decline compared to the original ILDC for CJPE study, even after implementing various weightage schemes in our DELSumm algorithm. Additionally, using only facts for legal judgment prediction with different transformer models yielded results inferior to the state-of-the-art outcomes reported in the "ILDC for CJPE" study.
\end{abstract}




\maketitle


\section{Introduction}
The motivation for this extended work is driven by the objective of making early predictions about legal case outcomes based solely on available facts, a common situation in real-world scenarios. This contrasts with the approach of the "ILDC for CJPE" project \cite{malik-etal-2021-ildc}, which utilized complete judgments, excluding the final decision. The "ILDC for CJPE" study was pivotal in advancing the field of legal document analysis. It introduced a comprehensive corpus of Indian legal documents, enabling sophisticated court judgment prediction and explanation. By analyzing a vast array of case documents, the study provided insights into the intricacies of legal reasoning and decision-making processes in the Indian judicial system. The project's methodologies and findings laid the groundwork for further exploration in the domain of legal AI, particularly in the context of judgment prediction based on varied case components.

Building on this foundation, our current focus is on scenarios where only case facts are available, without the benefit of lawyer arguments, statutes, lower court rulings, precedents, and other enriching context. Looking ahead, our plan includes refining model architectures, integrating Large Language Models (LLMs), and expanding our datasets with richer annotations. These efforts aim to enhance the model's ability to accurately predict outcomes based solely on case facts. This approach seeks to improve legal literacy, transparency, and inclusivity. By delivering predictions in the early stages of legal proceedings, our extended approach not only advances the field of legal AI but also contributes significantly to the public understanding of legal processes.

\section{Related Work}
The foundational paper "ILDC for CJPE" \cite{malik-etal-2021-ildc} inspired this extended study, supported by related works like "DELSumm \cite{bhattacharya-icail2021}," "Identification of Rhetorical Roles of Sentences in Indian Legal Judgments \cite{ghosh2019identification}," and the research on semantic segmentation of legal documents \cite{malik2021semantic}. These works provide a framework for our methodological design. A shared task~\cite{nigam2023nonet} on "ILDC for CJPE" has been undertaken. 

One of the foundational papers in the area of legal judgment prediction is ~\cite{medvedeva2020using}. This paper explores the use of machine learning techniques to predict the outcomes of cases in the European Court of Human Rights.
Hindi legal Document Corpus~\cite{kapoor-etal-2022-hldc} is a dataset provided for the purpose of Bail prediction in Hindi Language. The paper aims at predicting whether the appellant would get bail or not as a binary classification task by using a battery of models. 
Chinese Legal dataset~\cite{xiao2018cail2018} for Judgement Prediction is a dataset released in the form of a challenge in Chinese Language which aims to attempt the problem of Legal judgment prediction. The challenge is aimed at using transformer-based models and other LLMs to attempt the critical problem of legal judgment prediction in the Chinese language. The development of legal NLP datasets and other challenges has been crucial for advancing research in this area.

\section{Problem Statement}
\subsection{Proposed task}
The task of developing a Legal Judgment Prediction System centers around employing machine learning and natural language processing (NLP) techniques to predict the outcomes of legal cases based on the facts presented. This specialized application focuses on forecasting whether a case will be ruled in favor of the appellant/petitioner or the respondent.

\subsection{Task Description for Court Judgment Prediction and Explanation (CJPE)}
This task is designed to predict the final verdict of a case, considering all the facts and arguments, and to provide a rationale for the predicted outcome. The CJPE task involves two key components:

\begin{enumerate}
    \item \textbf{Prediction:} This sub-task requires predicting the outcome of a case (denoted as $y \in {0, 1}$), where `1' represents a verdict in favor of the appellant/petitioner. This is done by analyzing the case proceedings and applying predictive modeling.
    \item \textbf{Explanation:} Following the prediction, the system is tasked with explaining the verdict by identifying crucial sentences or phrases from the case proceedings that justify the decision. This explanation phase is crucial, as it provides insight into the reasoning behind the prediction. Notably, these explanations are not part of the training data, emphasizing the model's ability to generate justifications independently. This approach is adopted because obtaining annotated explanations for training is challenging.
\end{enumerate}

Overall, the CJPE task, particularly in the context of the Supreme Court of India (SCI), involves both predicting the court's decision regarding an appeal and elucidating the reasoning behind this prediction, enhancing the transparency and understandability of the legal decision-making process.

\section{Datasets}
\subsection{Existing Datasets}
The primary dataset employed is the ILDC for CJPE. This corpus is enhanced with retrieved facts and is used to test our two variations.

\subsection{Data Collection and Preprocessing}
To facilitate the task of fact-based judgment prediction, we plan to assemble a comprehensive dataset of openly available Indian legal statutes, including the constitution, sections, articles, acts, and related legal documents. Additionally, we will compile a database of Supreme Court Indian cases to serve as a foundational resource for our predictive model.

\subsubsection{Incorporating External Knowledge}
We plan to enhance our dataset by appending detailed descriptions of these statutes to relevant cases, providing additional context and information where specified or applicable. This enrichment will contribute to a more informative and comprehensive dataset.

\subsubsection{Data Preprocessing}
Effective data preprocessing is essential to ensure the quality and genericity of the dataset. To achieve this, we will employ the following preprocessing steps:

\textit{NER Tag Masking:} We will apply Named Entity Recognition (NER) tag masking to sensitive information such as the names of defendants, petitioners, judges, locations, organization names, and other specific entities. This masking process will enhance the genericity of the dataset while preserving its informative content.
By meticulously curating and preprocessing the dataset in this manner, we aim to create a robust foundation for our fact-based judgment prediction task. These efforts will ensure that the dataset is well-suited to support the development of accurate and generalizable predictive models.

\section{Planned Methodology}
To perform our task, we employ Hierarchical BiLSTM-CRF classifier \cite{ghosh2019identification} for extracting rhetorical roles, and to retrieve similar past cases, we utilize tools like "U-CREAT" \cite{joshi2023u} and "nigam@COLIEE-22"\cite{nigam2022nigam}. The algorithmic design is geared towards extracting key features, such as facts, statutes, ratio of decision, etc. We also implement different rhetorical role weights for our two variations of the DELSumm algorithm. The DELSumm algorithm gives different weights to the various segments or rhetorical roles in a legal judgment thereby creating summaries suitably tailored for the purpose of legal judgment summarization. Since transformer models have length limitations and legal judgments are extremely long in length, feeding the legal judgments into the transformers is quite difficult. Hence we need to 
summarize the legal judgments before feeding the legal judgment into the transformer-based models. DELSumm is an extractive summarization method specifically tailored for Indian legal Judgements which has been shown to effectively summarize Indian legal Judgments. So we use the DELSumm algorithm to summarize Indian Legal judgments to feed into the transformer-based models (Bert, LegalBert, InlegalBert, etc.) for the purpose of legal judgment prediction.

Another technique that we intend to try is to augment the dataset with various rhetorical roles from previous similar cases, including facts, statutes, the ratio of the decision, ruling by the present court, and more. Since previous similar cases will follow similar decision patterns corresponding to a particular legal case. Hence taking cues from a previous similar legal case can improve the quality of results in a legal judgment prediction task.

Additionally, we plan to include a comprehensive description of all statutes related to the case, addressing a gap where trained models often lack details about the statutes mentioned in legal cases.

\subsection{Pretrained Models and Specific Pretraining}
\subsubsection{Extract Rhetorical Roles}
A rhetorical role represents the semantic meaning of a sentence in a judgement~\cite{ghosh2019identification}.
To extract the rhetorical roles of the case, we will draw upon existing techniques and tools such as "DELSumm: Incorporating Domain Knowledge for Extractive Summarization of Legal Case Documents" \cite{bhattacharya-icail2021}, "Identification of Rhetorical Roles of Sentences in Indian Legal Judgments" \cite{ghosh2019identification}, "Corpus for Automatic Structuring of Legal Documents" \cite{kalamkar2022corpus}, and "Semantic Segmentation of Legal Documents via Rhetorical Roles" \cite{malik2021semantic}.

DELSumm, an unsupervised extractive algorithm, is a vital component of our methodology. It operates specifically in the legal domain, summarizing case documents based on rhetorical roles assigned to each sentence (e.g., Facts, Issue, Final Judgement, Arguments), and has guidelines for the desired summary length and content. This tool employs Integer Linear Programming (ILP)-based optimization for generating summaries.

\subsubsection{Prior Case Retrieval}
To identify previous similar cases, we will utilize techniques like "U-CREAT: Unsupervised Case Retrieval using Events extrAcTion" \cite{joshi2023u} and "nigam@COLIEE-22: Legal Case Retrieval and Entailment Using Cascading of Lexical and Semantic-Based Models" \cite{nigam2022nigam}.

\subsubsection{Judgment Prediction Models}
For fact-based judgment prediction, we employ a diverse set of models, each tailored to specific aspects of the task. These models include BERT \cite{devlin2018bert}, LegalBERT \cite{chalkidis2020legal}, InlegalBert \cite{paul-2022-pretraining}, InCaseLaw \cite{paul-2022-pretraining}, Roberta \cite{liu2019roberta}, and XLNet \cite{yang2019xlnet}. These models are trained on both general and legal-specific datasets to ensure comprehensive coverage. Given the inherent constraints of input token limits (512 tokens), we implement a hierarchical attention architecture with a sliding window approach. This approach facilitates the cascading of semantic information from one chunk to the next, enabling a thorough analysis of complex legal case documents.

\subsection{Facts Extraction Training Approach}
To focus specifically on extracting the facts of the case, we adopted a specialized training approach for the Hierarchical BiLSTM-CRF model. Instead of treating the task as a multi-class classification problem with multiple rhetorical roles, we transformed it into a binary classification task: distinguishing between facts and non-fact statements.

In this approach, we consider all rhetorical roles other than facts, such as issue, statute, Ratio of the decision, precedent, and arguments, as non-Fact statements. By simplifying the problem into a binary classification scenario, our model learns to discern the crucial Facts of the case from the remaining content.

This binary classification strategy was chosen to streamline the model's focus on the most critical information for judgment prediction. By doing so, we aim to enhance the model's accuracy and effectiveness in identifying and extracting the facts that play a pivotal role in the early phase of a legal case.

This approach is aligned with our goal of facilitating fact-based judgment prediction and providing valuable insights into legal outcomes based on factual information within the case documents.

\subsection{Evaluation and Validation}
Our approach to judgment prediction and explanation in legal documents involves a robust evaluation and validation process to ensure the accuracy, reliability, and efficacy of the generated results. Given the critical nature of legal information and its potential impact on decision-making, it is imperative to employ rigorous assessment methods.

\subsubsection{Performance Metrics} 
In assessing the quality of our judgment prediction and explanation models, we employ a range of performance metrics tailored specifically for the legal domain. These metrics include:

\begin{itemize}
    \item Precision: Precision measures the accuracy of positive predictions, reflecting the ratio of correctly predicted judgments to the total positive predictions. In the legal context, precision is crucial as it signifies the model's ability to provide accurate information about case outcomes.
    \item Recall: Recall evaluates the model's ability to identify all relevant positive cases, representing the ratio of correctly predicted judgments to the total actual positive cases. High recall is essential in ensuring comprehensive coverage of case outcomes.
    \item F1-Score: The F1-Score balances precision and recall, providing a harmonic mean that considers both false positives and false negatives. This metric is particularly valuable in legal applications, as it gauges the overall predictive performance.
\end{itemize}

\subsection{User-Friendly Interface}
Recognizing the diverse user base of our judgment prediction, we strongly emphasize designing an intuitive and accessible user interface. Our goal is to make legal information readily available and comprehensible to both legal practitioners and the general public, promoting transparency, inclusivity, and legal literacy.

\section{Results and Analysis}
Our preliminary experimentation with the ILDC for CJPE dataset for judgment prediction, after retrieving facts statements, yielded unsatisfactory results. We attempted various state-of-the-art models, including BERT, XLNet, Roberta, LegalBERT, InLegalBERT, and InCaseLaw, along with hierarchical forms with attention networks to accommodate the document's entirety, considering the input token constraints of 512 tokens. 

We also explored two variations of DELSumm: a standard weightage scheme (variation 1) and a modified weightage scheme (variation 2) for the summarized context of different rhetorical roles. In variation 1, the importance given to each rhetorical role was as follows:

\begin{itemize}
    \item 128 for 'Final Judgment'
    \item 64 for 'Issue'
    \item 32 for 'Fact'
    \item 8 for 'Statute,' 'Ratio of the decision,' and 'Precedent'
    \item 2 for 'Argument'
\end{itemize}
The first variant is the default parameter setting for DELSumm where we use default segment weights provided in ~\cite{bhattacharya-icail2021}.
In variation 2, the importance was as follows:
\begin{itemize}
    \item 128 for 'Final Judgment'
    \item 64 for 'Ratio of the decision'
    \item 32 for 'Argument'
    \item 8 for 'Statute,' 'Issue,' and 'Facts'
    \item 2 for 'Precedent'
\end{itemize}
The second variation is based on our specific needs for providing greater segment weights for the Final Judgement and Ratio of the decision which is extremely important for judgment prediction. Followed by these roles we have Argument, Statute, Issue, Facts, and Precedent. Arguments are also important for deciding the final judgment of a legal case followed by Facts, Statutes, and Precedent of the legal case. 

We observed that variation 2 which is specifically tailored to our own needs performs better than variation 1 which was the default parameter setting for DELSumm.
The kind of information that the summarizer chooses to be part of the generated extractive summary plays a critical role in deciding the quality of judgment prediction.
Also, we summarized only the facts of a legal judgment using DELsumm as well the facts+Ruling by the lower court and then fed those into the transformer models.


\begin{table}[h!]
\small
\caption{Results of XLNET transformer model on ILDC multi dataset based on the different variations of the DELSumm summarization algorithm. Technique 1: the sliding window of 410 tokens with an overlap of 100 tokens, Technique 2: last 510 tokens}
    \label{tab:t3}
    \centering
\scalebox{0.77}{
\begin{tabular}{c|c|c|c|c}
\hline
\textbf{\begin{tabular}[c]{@{}c@{}}Transformer model\\ \end{tabular}} & \textbf{Precision} & \textbf{Recall} & \textbf{F1-Score} & \textbf{Technique} \\ \hline

Hierarchical XLNET(state of the art)~\cite{malik-etal-2021-ildc}  & 0.7780 & 0.7778 & 0.7779 & 1 \\ \hline

XLNET (variation 1) & 0.5619 & 0.5282 & 0.5445 & 2 \\ \hline

XLNET (variation 2) & 0.7076 & 0.6932 & 0.7003 & 2  \\ \hline

XLNET (only facts) & 0.6016 & 0.6016 & 0.6016 & 2  \\ \hline

XLNET (facts+Ruling by lower court) & 0.6168 & 0.6167 & 0.6167 & 2  \\ \hline

XLNET (variation 1) & 0.6479 & 0.6298 & 0.6387 & 1   \\ \hline

XLNET (variation 2) & 0.7303 & 0.7153 & 0.7227 & 1   \\ \hline

XLNET (only facts) & 0.5963 & 0.5624 & 0.5788 & 1   \\ \hline

XLNET (facts+Ruling by lower court) & 0.5907 & 0.5463 & 0.5674 & 1  \\ \hline


    \end{tabular}
}

\end{table}



\begin{table}[h!]
\small
\caption{Results of Different transformer models with first 510 tokens and Hierarchical transformer models on the ILDC multi dataset. Technique 1: sliding window of 410 tokens with overlap of 100 tokens, Technique 2: last 510 tokens, Technique 3: first 510 tokens}
    \label{tab:t3}
    \centering
\scalebox{0.77}{
\begin{tabular}{c|c|c|c|c}
\hline
\textbf{\begin{tabular}[c]{@{}c@{}}Transformer model\\ \end{tabular}} & \textbf{Precision} & \textbf{Recall} & \textbf{F1-Score} & \textbf{Technique} \\ \hline

Hierarchical XLNET(state of the art)~\cite{malik-etal-2021-ildc}  & 0.7780 & 0.7778 & 0.7779 & 1 \\ \hline

XLNET-large & 0.5866 & 0.5452 & 0.5651 & 3 \\ \hline

XLNET-large & 0.6024 & 0.5736 & 0.5876 & 1  \\ \hline

InLegalBert & 0.5907 & 0.5371 & 0.5626 & 2  \\ \hline

InlegalBert & 0.5934 & 0.5802 & 0.5868 & 1   \\ \hline

InCaselawBert & 0.5787 & 0.5522 & 0.5651 & 3  \\ \hline

IncaselawBert & 0.5797 & 0.5612 & 0.5703 & 1   \\ \hline

Bert-base & 0.5565 & 0.5530 & 0.5547 & 2  \\ \hline

Bert-base & 0.5439 & 0.5438 & 0.5438 & 1  \\ \hline


    \end{tabular}
}

\end{table}

Despite these efforts, the results were considerably poorer than the original "ILDC for CJPE" paper~\cite{malik-etal-2021-ildc} even after using different weightage schemes in our DELSumm algorithm.

Also using only the facts of a legal judgment with different transformer models has given lower results than the state-of-the-art results presented in "ILDC for CJPE"~\cite{malik-etal-2021-ildc}.


\section{Task Complexity}
Predicting judgments based solely on the facts presented in legal documents introduces a significantly higher level of complexity compared to approaches that consider the entire case, as demonstrated in the ILDC for the CJPE paper. The prior results from that work already indicated a substantial gap between the model predictions and expert scores, even when utilizing the full case content except for the final decision.

This heightened complexity arises from several factors:

\begin{enumerate}

    \item \textbf{Limited Information:} Relying solely on case facts limits the data available to models, making them heavily dependent on the quality and thoroughness of fact extraction. We aim to enhance this by incorporating data from similar past cases, identified through a similarity detection process, into the current case facts.
    
    
    \item \textbf{Error Propagation:} Fact extraction methods vary and can introduce errors, complicating judgment prediction. To mitigate this, we propose obtaining factual data directly from law firms, ensuring access to accurate and complete case facts for better prediction outcomes.
    

    \item \textbf{Reduced Context} Predictions based only on facts miss the wider context of legal disputes, such as lawyer arguments, statutes, lower court rulings, and precedents. To address this, we plan to enrich the case facts with information from similar past judgments, thereby offering a more nuanced understanding and improving prediction quality.
    

    \item \textbf{Expert Discrepancies:} The variation between model predictions and expert opinions, even with more complete case data, highlights the inherent complexity. Fact-based predictions intensify this, as even panels of judges often differ in their verdicts, with majority opinions guiding the final decision.
\end{enumerate}
Addressing these complexities in fact-based judgment prediction requires innovative approaches in data collection, preprocessing, feature engineering, and modeling to achieve reliable and accurate results.

\section{Conclusion and Future Work}
While our variations did not significantly outperform the original model, they shed light on the challenges and complexities of predicting legal judgments based solely on facts or combined with RLCs. Future research should consider integrating richer contextual information, combining structured and unstructured data, and exploring more advanced neural architectures to enhance prediction accuracy.

Future work will also include the use of powerful architecture like GPT-4 and GPT-3.5 which gave decent reasoning abilities in a Zero-Shot mode as well as in a fine-tuned manner. Models like GPT have a large knowledge bank including on legal matters like statutes, and precedents of legal cases which are publically available because models like GPT have been trained on huge amounts of corpus. So, the use of architecture like GPT can prove fruitful.

Moreover, the expansion of our dataset is on the horizon. Increasing the volume and diversity of data is crucial as it equips the model with a broader learning spectrum, thereby enhancing its predictive abilities. This expansion will include the meticulous collection and preprocessing of Indian legal documents, incorporating external knowledge like detailed statute descriptions into relevant cases for contextual richness. Our data preprocessing will focus on maintaining the quality and genericity of the dataset, employing techniques like Named Entity Recognition (NER) tag masking to sensitive information, ensuring both privacy and informativeness.

Also enhancing the amount of data can be a possible future direction in this research as increasing the amount of data helps the model learn better. Also, the use of Encoder-Decoder architectures for solving problems on Legal Judgement prediction can be attempted.
The exploration of this space promises significant practical and academic value in the years to come.

\bibliographystyle{ACM-Reference-Format}
\bibliography{sample-base}


\begin{thebibliography}{16}


\ifx \showCODEN    \undefined \def \showCODEN     #1{\unskip}     \fi
\ifx \showDOI      \undefined \def \showDOI       #1{#1}\fi
\ifx \showISBNx    \undefined \def \showISBNx     #1{\unskip}     \fi
\ifx \showISBNxiii \undefined \def \showISBNxiii  #1{\unskip}     \fi
\ifx \showISSN     \undefined \def \showISSN      #1{\unskip}     \fi
\ifx \showLCCN     \undefined \def \showLCCN      #1{\unskip}     \fi
\ifx \shownote     \undefined \def \shownote      #1{#1}          \fi
\ifx \showarticletitle \undefined \def \showarticletitle #1{#1}   \fi
\ifx \showURL      \undefined \def \showURL       {\relax}        \fi
\providecommand\bibfield[2]{#2}
\providecommand\bibinfo[2]{#2}
\providecommand\natexlab[1]{#1}
\providecommand\showeprint[2][]{arXiv:#2}

\bibitem[Bhattacharya et~al\mbox{.}(2021)]%
        {bhattacharya-icail2021}
\bibfield{author}{\bibinfo{person}{Paheli Bhattacharya}, \bibinfo{person}{Soham Poddar}, \bibinfo{person}{Koustav Rudra}, \bibinfo{person}{Kripabandhu Ghosh}, {and} \bibinfo{person}{Saptarshi Ghosh}.} \bibinfo{year}{2021}\natexlab{}.
\newblock \showarticletitle{{Incorporating Domain Knowledge for Extractive Summarization of Legal Case Documents}}. In \bibinfo{booktitle}{\emph{{Proceedings of the 18th International Conference on Artificial Intelligence and Law (ICAIL)}}}.
\newblock


\bibitem[Chalkidis et~al\mbox{.}(2020)]%
        {chalkidis2020legal}
\bibfield{author}{\bibinfo{person}{Ilias Chalkidis}, \bibinfo{person}{Manos Fergadiotis}, \bibinfo{person}{Prodromos Malakasiotis}, \bibinfo{person}{Nikolaos Aletras}, {and} \bibinfo{person}{Ion Androutsopoulos}.} \bibinfo{year}{2020}\natexlab{}.
\newblock \showarticletitle{LEGAL-BERT: The muppets straight out of law school}.
\newblock \bibinfo{journal}{\emph{arXiv preprint arXiv:2010.02559}} (\bibinfo{year}{2020}).
\newblock


\bibitem[Devlin et~al\mbox{.}(2018)]%
        {devlin2018bert}
\bibfield{author}{\bibinfo{person}{Jacob Devlin}, \bibinfo{person}{Ming-Wei Chang}, \bibinfo{person}{Kenton Lee}, {and} \bibinfo{person}{Kristina Toutanova}.} \bibinfo{year}{2018}\natexlab{}.
\newblock \showarticletitle{Bert: Pre-training of deep bidirectional transformers for language understanding}.
\newblock \bibinfo{journal}{\emph{arXiv preprint arXiv:1810.04805}} (\bibinfo{year}{2018}).
\newblock


\bibitem[Ghosh and Wyner(2019)]%
        {ghosh2019identification}
\bibfield{author}{\bibinfo{person}{Saptarshi Ghosh} {and} \bibinfo{person}{Adam Wyner}.} \bibinfo{year}{2019}\natexlab{}.
\newblock \showarticletitle{Identification of rhetorical roles of sentences in indian legal judgments}.
\newblock \bibinfo{journal}{\emph{Legal Knowledge and Information Systems: JURIX}} (\bibinfo{year}{2019}), \bibinfo{pages}{3}.
\newblock


\bibitem[Joshi et~al\mbox{.}(2023)]%
        {joshi2023u}
\bibfield{author}{\bibinfo{person}{Abhinav Joshi}, \bibinfo{person}{Akshat Sharma}, \bibinfo{person}{Sai~Kiran Tanikella}, {and} \bibinfo{person}{Ashutosh Modi}.} \bibinfo{year}{2023}\natexlab{}.
\newblock \showarticletitle{U-CREAT: Unsupervised Case Retrieval using Events extrAcTion}.
\newblock \bibinfo{journal}{\emph{arXiv preprint arXiv:2307.05260}} (\bibinfo{year}{2023}).
\newblock


\bibitem[Kalamkar et~al\mbox{.}(2022)]%
        {kalamkar2022corpus}
\bibfield{author}{\bibinfo{person}{Prathamesh Kalamkar}, \bibinfo{person}{Aman Tiwari}, \bibinfo{person}{Astha Agarwal}, \bibinfo{person}{Saurabh Karn}, \bibinfo{person}{Smita Gupta}, \bibinfo{person}{Vivek Raghavan}, {and} \bibinfo{person}{Ashutosh Modi}.} \bibinfo{year}{2022}\natexlab{}.
\newblock \showarticletitle{Corpus for automatic structuring of legal documents}.
\newblock \bibinfo{journal}{\emph{arXiv preprint arXiv:2201.13125}} (\bibinfo{year}{2022}).
\newblock


\bibitem[Kapoor et~al\mbox{.}(2022)]%
        {kapoor-etal-2022-hldc}
\bibfield{author}{\bibinfo{person}{Arnav Kapoor}, \bibinfo{person}{Mudit Dhawan}, \bibinfo{person}{Anmol Goel}, \bibinfo{person}{Arjun T~H}, \bibinfo{person}{Akshala Bhatnagar}, \bibinfo{person}{Vibhu Agrawal}, \bibinfo{person}{Amul Agrawal}, \bibinfo{person}{Arnab Bhattacharya}, \bibinfo{person}{Ponnurangam Kumaraguru}, {and} \bibinfo{person}{Ashutosh Modi}.} \bibinfo{year}{2022}\natexlab{}.
\newblock \showarticletitle{{HLDC}: {H}indi Legal Documents Corpus}. In \bibinfo{booktitle}{\emph{Findings of the Association for Computational Linguistics: ACL 2022}}. \bibinfo{publisher}{Association for Computational Linguistics}, \bibinfo{address}{Dublin, Ireland}, \bibinfo{pages}{3521--3536}.
\newblock
\urldef\tempurl%
\url{https://doi.org/10.18653/v1/2022.findings-acl.278}
\showDOI{\tempurl}


\bibitem[Liu et~al\mbox{.}(2019)]%
        {liu2019roberta}
\bibfield{author}{\bibinfo{person}{Yinhan Liu}, \bibinfo{person}{Myle Ott}, \bibinfo{person}{Naman Goyal}, \bibinfo{person}{Jingfei Du}, \bibinfo{person}{Mandar Joshi}, \bibinfo{person}{Danqi Chen}, \bibinfo{person}{Omer Levy}, \bibinfo{person}{Mike Lewis}, \bibinfo{person}{Luke Zettlemoyer}, {and} \bibinfo{person}{Veselin Stoyanov}.} \bibinfo{year}{2019}\natexlab{}.
\newblock \showarticletitle{Roberta: A robustly optimized bert pretraining approach}.
\newblock \bibinfo{journal}{\emph{arXiv preprint arXiv:1907.11692}} (\bibinfo{year}{2019}).
\newblock


\bibitem[Malik et~al\mbox{.}(2021a)]%
        {malik2021semantic}
\bibfield{author}{\bibinfo{person}{Vijit Malik}, \bibinfo{person}{Rishabh Sanjay}, \bibinfo{person}{Shouvik~Kumar Guha}, \bibinfo{person}{Angshuman Hazarika}, \bibinfo{person}{Shubham Nigam}, \bibinfo{person}{Arnab Bhattacharya}, {and} \bibinfo{person}{Ashutosh Modi}.} \bibinfo{year}{2021}\natexlab{a}.
\newblock \showarticletitle{Semantic segmentation of legal documents via rhetorical roles}.
\newblock \bibinfo{journal}{\emph{arXiv preprint arXiv:2112.01836}} (\bibinfo{year}{2021}).
\newblock


\bibitem[Malik et~al\mbox{.}(2021b)]%
        {malik-etal-2021-ildc}
\bibfield{author}{\bibinfo{person}{Vijit Malik}, \bibinfo{person}{Rishabh Sanjay}, \bibinfo{person}{Shubham~Kumar Nigam}, \bibinfo{person}{Kripabandhu Ghosh}, \bibinfo{person}{Shouvik~Kumar Guha}, \bibinfo{person}{Arnab Bhattacharya}, {and} \bibinfo{person}{Ashutosh Modi}.} \bibinfo{year}{2021}\natexlab{b}.
\newblock \showarticletitle{{ILDC} for {CJPE}: {I}ndian Legal Documents Corpus for Court Judgment Prediction and Explanation}. In \bibinfo{booktitle}{\emph{Proceedings of the 59th Annual Meeting of the Association for Computational Linguistics and the 11th International Joint Conference on Natural Language Processing (Volume 1: Long Papers)}}. \bibinfo{publisher}{Association for Computational Linguistics}, \bibinfo{address}{Online}, \bibinfo{pages}{4046--4062}.
\newblock
\urldef\tempurl%
\url{https://doi.org/10.18653/v1/2021.acl-long.313}
\showDOI{\tempurl}


\bibitem[Medvedeva et~al\mbox{.}(2020)]%
        {medvedeva2020using}
\bibfield{author}{\bibinfo{person}{Masha Medvedeva}, \bibinfo{person}{Michel Vols}, {and} \bibinfo{person}{Martijn Wieling}.} \bibinfo{year}{2020}\natexlab{}.
\newblock \showarticletitle{Using machine learning to predict decisions of the European Court of Human Rights}.
\newblock \bibinfo{journal}{\emph{Artificial Intelligence and Law}}  \bibinfo{volume}{28} (\bibinfo{year}{2020}), \bibinfo{pages}{237--266}.
\newblock


\bibitem[Nigam et~al\mbox{.}(2023)]%
        {nigam2023nonet}
\bibfield{author}{\bibinfo{person}{Shubham~Kumar Nigam}, \bibinfo{person}{Aniket Deroy}, \bibinfo{person}{Noel Shallum}, \bibinfo{person}{Ayush~Kumar Mishra}, \bibinfo{person}{Anup Roy}, \bibinfo{person}{Shubham~Kumar Mishra}, \bibinfo{person}{Arnab Bhattacharya}, \bibinfo{person}{Saptarshi Ghosh}, {and} \bibinfo{person}{Kripabandhu Ghosh}.} \bibinfo{year}{2023}\natexlab{}.
\newblock \showarticletitle{Nonet at SemEval-2023 Task 6: Methodologies for Legal Evaluation}. In \bibinfo{booktitle}{\emph{Proceedings of the The 17th International Workshop on Semantic Evaluation (SemEval-2023)}}. \bibinfo{pages}{1293--1303}.
\newblock


\bibitem[Nigam et~al\mbox{.}(2022)]%
        {nigam2022nigam}
\bibfield{author}{\bibinfo{person}{Shubham~Kumar Nigam}, \bibinfo{person}{Navansh Goel}, {and} \bibinfo{person}{Arnab Bhattacharya}.} \bibinfo{year}{2022}\natexlab{}.
\newblock \showarticletitle{nigam@ COLIEE-22: Legal Case Retrieval and Entailment using Cascading of Lexical and Semantic-based models}. In \bibinfo{booktitle}{\emph{JSAI International Symposium on Artificial Intelligence}}. Springer, \bibinfo{pages}{96--108}.
\newblock


\bibitem[Paul et~al\mbox{.}(2023)]%
        {paul-2022-pretraining}
\bibfield{author}{\bibinfo{person}{Shounak Paul}, \bibinfo{person}{Arpan Mandal}, \bibinfo{person}{Pawan Goyal}, {and} \bibinfo{person}{Saptarshi Ghosh}.} \bibinfo{year}{2023}\natexlab{}.
\newblock \showarticletitle{Pre-trained Language Models for the Legal Domain: A Case Study on Indian Law}. In \bibinfo{booktitle}{\emph{Proceedings of 19th International Conference on Artificial Intelligence and Law - ICAIL 2023}}.
\newblock
\urldef\tempurl%
\url{https://arxiv.org/abs/2209.06049}
\showURL{%
\tempurl}


\bibitem[Xiao et~al\mbox{.}(2018)]%
        {xiao2018cail2018}
\bibfield{author}{\bibinfo{person}{Chaojun Xiao}, \bibinfo{person}{Haoxi Zhong}, \bibinfo{person}{Zhipeng Guo}, \bibinfo{person}{Cunchao Tu}, \bibinfo{person}{Zhiyuan Liu}, \bibinfo{person}{Maosong Sun}, \bibinfo{person}{Yansong Feng}, \bibinfo{person}{Xianpei Han}, \bibinfo{person}{Zhen Hu}, \bibinfo{person}{Heng Wang}, {et~al\mbox{.}}} \bibinfo{year}{2018}\natexlab{}.
\newblock \showarticletitle{Cail2018: A large-scale legal dataset for judgment prediction}.
\newblock \bibinfo{journal}{\emph{arXiv preprint arXiv:1807.02478}} (\bibinfo{year}{2018}).
\newblock


\bibitem[Yang et~al\mbox{.}(2019)]%
        {yang2019xlnet}
\bibfield{author}{\bibinfo{person}{Zhilin Yang}, \bibinfo{person}{Zihang Dai}, \bibinfo{person}{Yiming Yang}, \bibinfo{person}{Jaime Carbonell}, \bibinfo{person}{Russ~R Salakhutdinov}, {and} \bibinfo{person}{Quoc~V Le}.} \bibinfo{year}{2019}\natexlab{}.
\newblock \showarticletitle{Xlnet: Generalized autoregressive pretraining for language understanding}.
\newblock \bibinfo{journal}{\emph{Advances in neural information processing systems}}  \bibinfo{volume}{32} (\bibinfo{year}{2019}).
\newblock


\end{thebibliography}

\end{document}